\documentclass{article}

 \usepackage[preprint]{neurips_2026}

\usepackage[utf8]{inputenc} 
\usepackage[T1]{fontenc}    
\usepackage{hyperref}       
\usepackage{url}            
\usepackage{booktabs}       
\usepackage{cellspace}
\usepackage{amsfonts}       
\usepackage{nicefrac}       
\usepackage{microtype}      
\usepackage{xcolor}         
\usepackage{graphicx}
\usepackage{xspace}
\usepackage{listings}
\usepackage{multirow}
\usepackage[table]{xcolor}
\usepackage{amsmath}
\usepackage{booktabs}
\definecolor{MementoLight}{RGB}{232,244,255}
\usepackage[most]{tcolorbox}
\newcommand{\memcell}[1]{\cellcolor{violet!10}{#1}}
\definecolor{eclipseStrings}{RGB}{42,0,255}
\definecolor{eclipseKeywords}{RGB}{127,0,85}
\definecolor{jsonPropertyName}{RGB}{0,0,128}
\definecolor{jsonValue}{RGB}{0,128,0}
\definecolor{PanelBlue}{HTML}{E6EFF6}
\definecolor{PanelTeal}{HTML}{E4F0EE}
\definecolor{HeaderBlue}{HTML}{C8D8E7}
\definecolor{HeaderTeal}{HTML}{C7DDD9}
\definecolor{MemoryBlue}{HTML}{DCE8F3}
\definecolor{MemoryTeal}{HTML}{DDEBE8}
\definecolor{RuleGray}{HTML}{B8C6D1}
\definecolor{my_green}{RGB}{51,102,0}
\definecolor{my_red}{RGB}{204, 0, 0}
\definecolor{my_gray}{RGB}{102,102,102}

\lstdefinelanguage{json}{
    basicstyle=\small\ttfamily,
    numbers=left,
    numberstyle=\tiny\color{gray},
    stepnumber=1,
    numbersep=8pt,
    showstringspaces=false,
    breaklines=true,
    frame=lines,
    backgroundcolor=\color{gray!5},
    string=[s]{"}{"},
    stringstyle=\color{jsonValue},
    comment=[l]{//},
    morecomment=[s]{/*}{*/},
    commentstyle=\color{gray},
    literate=
     *{0}{{{\color{eclipseKeywords}0}}}{1}
      {1}{{{\color{eclipseKeywords}1}}}{1}
      {2}{{{\color{eclipseKeywords}2}}}{1}
      {3}{{{\color{eclipseKeywords}3}}}{1}
      {4}{{{\color{eclipseKeywords}4}}}{1}
      {5}{{{\color{eclipseKeywords}5}}}{1}
      {6}{{{\color{eclipseKeywords}6}}}{1}
      {7}{{{\color{eclipseKeywords}7}}}{1}
      {8}{{{\color{eclipseKeywords}8}}}{1}
      {9}{{{\color{eclipseKeywords}9}}}{1}
      {:}{{{\color{jsonPropertyName}{:}}}}{1}
      {,}{{{\color{jsonPropertyName}{,}}}}{1}
      {\{}{{{\color{jsonPropertyName}{\{}}}}{1}
      {\}}{{{\color{jsonPropertyName}{\}}}}}{1}
      {[}{{{\color{jsonPropertyName}{[}}}}{1}
      {]}{{{\color{jsonPropertyName}{]}}}}{1},
}

\title{MementoGUI: Learning Agentic Multimodal Memory Control for Long-Horizon GUI Agents}

\newcommand{\DATA}{\textsc{MementoGUI-Bench}\xspace}
\newcommand{\NAME}{\textsc{MementoGUI}\xspace}
\newcommand{\SYSTEM}{\textsc{MementoCore}\xspace}
\author{%
  Ziyun Zeng\textsuperscript{1,* }, Hang Hua\textsuperscript{2,*,\(\dagger\) }, Bocheng Zou\textsuperscript{3 }, Mu Cai\textsuperscript{3 }, Rogerio Feris\textsuperscript{2 }, Jiebo Luo\textsuperscript{1}\\
  \textsuperscript{1}University of Rochester, \textsuperscript{2}MIT-IBM Watson AI Lab,
  \textsuperscript{3}University of Wisconsin-Madison\\
  \texttt{ziyun.zeng@rochester.edu,hang.hua1@ibm.com,bochengz@cs.wisc.edu}\\
  \texttt{mucai@cs.wisc.edu,rsferis@us.ibm.com,jluo@cs.rochester.edu}\\
  \textsuperscript{\(\dagger\)} Project Lead,   \textsuperscript{*} Equal Contribution
}

\begin{document}

\maketitle

\begin{abstract}
Recent GUI agents have made substantial progress in visual grounding and action prediction, yet they remain brittle in long-horizon tasks that require maintaining task state across many interface transitions. Existing agents typically rely on raw history replay or text-only memory, which either overwhelms the model with redundant screenshots or discards localized visual evidence needed for future decisions. To address these limitations, we introduce \textbf{\NAME}, a plug-in agentic memory framework that equips MLLM-based GUI agents with \textbf{\SYSTEM}, a learned controller for online memory selection, compression, and retrieval. Rather than treating interaction history as a fixed context, \NAME formulates long-horizon GUI control as an online memory-control problem: working memory selectively preserves task-relevant interface events with textual summaries and ROI-level visual evidence, while episodic memory retrieves reusable past trajectories through learned relevance selection. \SYSTEM modularizes memory control into specialized operators for step processing, memory compression, episodic writing, and episodic selection, enabling plug-in memory augmentation without finetuning the GUI agent backbone. We further develop a scalable data curation pipeline that converts computer-use trajectories into memory-controller training data, introduce \textbf{\DATA} for evaluating long-horizon decision-making in GUI agents, and design MLLM-based metrics for semantic action matching, task progress, and memory consistency. Experiments on GUI-Odyssey, MM-Mind2Web, and \DATA show that \NAME consistently improves GUI agents over no-history, history-replay, and text-only memory baselines, with larger \SYSTEM backbones further strengthening memory-augmented GUI control. Resources available at \href{https://zzzmyyzeng.github.io/MementoGUI}{\textcolor{my_green}{\fontfamily{cmtt}\selectfont{zzzmyyzeng.github.io/MementoGUI}}}
\end{abstract}
    
\section{Introduction}
Recent advances in multimodal large language models (MLLMs)~\cite{bai2025qwen3,hua2025v2xum,liu2023visual,singh2025openai,Sun2025LatentCF,wang2025internvl3} have enabled agentic systems that perceive, reason, and act in complex visual environments~\cite{avogaro2026sparc,Hu2023PromptCapPI,hua2025finecaption,hua2024mmcomposition,Hua2025MMIGBenchTC,thrush2022winoground,yu2024promptfix,yu2025omnipaint}, alongside their growing success in complex scientific tasks~\cite{cao2024presto,tang2025medagentsbench,tang2025cellforge,zeng2026automated,zeng2025use}. Graphical user interface (GUI) control is a representative setting for such agents, requiring visually grounded actions over dynamic software interfaces. While recent GUI agents have improved single-step grounding and action prediction~\cite{deng2023mind2web,gou2024navigating,Hua2024FINEMATCHAF,lei2025grounding,zeng2025mira,zheng2024gpt}, long-horizon GUI control remains brittle \cite{koh2024visualwebarena,lu2025guiodyssey,rawles2023androidinthewild,xie2024osworld,zhou2023webarena}. Agents have to preserve task state across many interface transitions, where crucial evidence can be local, transient, or unavailable in later screenshots, such as a selected widget state, a temporary menu option, or an earlier instruction needed for a later decision. As trajectories grow longer, these missed cues accumulate, causing agents to forget constraints, lose track of progress, or repeat ineffective actions. This failure mode appears in both cross-app mobile environments~\cite{lu2025guiodyssey,rawles2023androidinthewild} and multimodal web settings~\cite{deng2023mind2web}, suggesting a fundamental paradigm shift in GUI agent design: the primary bottleneck is no longer single-step visual understanding, but rather the active management of long-term multimodal state.

Existing GUI agents often address long-horizon interaction through passive history conditioning~\cite{gao2025chain,wang2024mobile,xu2026mobile,xu2025retrieval}. 
However, longer histories or text-only memory representations do not necessarily provide decision-useful context, and may introduce redundant or distracting information. 
In long GUI trajectories, useful evidence is sparse and unevenly distributed: some past steps only reflect routine transitions, while others encode task constraints, completed subgoals, or localized visual cues that may no longer be visible in the current screenshot. 
This suggests that long-horizon GUI control is better viewed as a multimodal memory-control problem rather than a pure context-length problem. 
Effective agents should decide when to update memory, what to preserve, how to compress interaction history, and when to retrieve past evidence for future decisions.

To address this challenge, we introduce \textbf{\NAME}, a plug-in agentic multimodal memory-control framework for long-horizon GUI agents. \NAME augments a frozen GUI backbone with a learned memory controller rather than finetuning the action policy itself. The controller maintains memory at two complementary timescales: working memory for evolving in-task state and episodic memory for reusable experience from prior interactions. At each step, the controller transforms relevant interaction history into structured multimodal context, including concise event summaries and localized visual references. The frozen GUI backbone then predicts actions from the current screenshot with the memory context, turning interaction history from passive context replay into a decision-oriented control layer.

Trained with large-scale supervision automatically curated from computer-use trajectories, \NAME consistently improves frozen GUI backbones across GUI-Odyssey~\cite{lu2025guiodyssey}, Multimodal-Mind2Web~\cite{deng2023mind2web}, and our \DATA. Beyond standard GUI metrics, we further evaluate long-horizon behavior with memory-aware metrics that measure semantic action matching, task progress, and memory consistency. For example, on GUI-Odyssey with UI-Venus-1.5-8B, \NAME improves action matching from 54.58 to 68.32 and trajectory success from 1.29 to 3.57, outperforming no-history, history-replay, and text-only memory baselines. These results support our central hypothesis that learning to control multimodal memory is more effective than relying on longer raw interaction histories or text-only memory representations for long-horizon GUI agents. Our contributions are summarized as follows:

\begin{itemize}

    \item We propose \NAME, a plug-in online multimodal agent memory framework that reframes long-horizon GUI control from raw history conditioning to active memory management. \NAME augments frozen GUI backbones with a learned controller that actively manages working and episodic memory, enabling agents to preserve and retrieve decision-relevant multimodal state without finetuning the underlying GUI action model.

    \item We develop an automatic data curation pipeline from PSAI computer-use trajectories to provide scalable supervision for memory control. 
    The pipeline converts raw interactions into training signals for step processing, working-memory compression, episodic memory writing, and episodic memory selection, enabling \NAME to learn memory operations with minimal trajectory-level annotation.

    \item We introduce \DATA, a benchmark for memory-dependent long-horizon GUI decision making, together with memory-aware metrics for semantic action matching, task progress, and memory consistency. Experiments across mobile and web environments show that \NAME consistently improves frozen GUI backbones over strong no-history, raw-history, and text-only memory baselines.

\end{itemize}

\section{Related Work}
\label{sec:related_work}
\paragraph{Memory Systems for Autonomous Agents.}
Recent GUI-agent research has explored memory mechanisms beyond raw interaction history. 
MGA~\cite{cheng2025mga} and adaptive history modeling~\cite{wu2025auto} improve within-task state tracking by managing long GUI trajectories more compactly. 
For cross-task reuse, Chain-of-Experience~\cite{gao2025chain}, EchoTrail~\cite{li2025echotrail}, and HybridAgent~\cite{zhu2026hybrid} store past trajectories as reasoning chains, retrievable traces, or structured knowledge. 
Other computer-use agents accumulate reusable knowledge through online interaction, demonstrations, or self-improvement, including AppAgentX~\cite{jiang2025appagentx}, MobileGPT~\cite{lee2024mobilegpt}, ScaleCUA~\cite{liu2025scalecua}, UI-Explorer~\cite{xiao2026ui}, EvoCUA~\cite{xue2026evocua}, and AppAgent~\cite{zhang2025appagent}. More broadly, autonomous-agent memory has developed around memory streams~\cite{park2023generative}, verbal replay~\cite{shinn2023reflexion}, skill libraries~\cite{wang2023voyager}, and procedural memory~\cite{fang2025memp,wang2024agent}, as well as self-updating memory and retrieval-augmented refinement~\cite{tang2025chemagent,tang2025eigen}. 
Recent systems further study learned memory control~\cite{hu2025hiagent,yu2026agentic}, trainable memory operations~\cite{wang2026infmem,zhang2026memskill}, self-organizing memory frameworks~\cite{guo2026memfactory,xu2025mem}, decision-theoretic memory management~\cite{sun2025beyond}, and efficient compressed or parametric memory representations~\cite{borro2026memori,liu2026simplemem,lu2026locas}. 
Multimodal memory systems have also begun to store visual trajectories for open-world planning~\cite{li2024optimus,wang2024jarvis}, unify visual and episodic memory for video reasoning~\cite{yeo2025worldmm}, or distill multimodal experience into reusable programs and lifelong memory~\cite{chen2025telemem,liu2025memverse,sarch2024vlm}. 
However, they do not fully address long-horizon GUI control, where dense screenshot streams must be selectively compressed, localized visual state changes must be preserved, and memory retrieval must directly support action prediction.

\paragraph{Long-Horizon Challenges in GUI Agents.}
Recent vision-language models have substantially advanced GUI automation, from visual grounding~\cite{cheng2024seeclick,hong2024cogagent,lin2025showui,huang2025dave} to cross-platform foundation action models~\cite{agashe2025agent,qin2025ui,wu2024atlas,huang2025building}.
Recent technical reports and open-source systems, including MAI-UI~\cite{zhou2025mai}, GUI-Owl-1.5~\cite{xu2026mobile}, Step-GUI~\cite{yan2025step}, and UI-Venus-1.5~\cite{gao2026ui}, further improve GUI grounding and navigation across desktop, web, and mobile settings.
Complementary efforts further improve efficiency through adaptive perception~\cite{mehrotra2025ishift}, compositional planning~\cite{agashe2025agent}, and systematic skill acquisition via exploration~\cite{liu2026osexpert,sun2025seagent}.
Yet long-horizon tasks remain a dominant failure mode: on benchmarks such as OSWorld~\cite{xie2024osworld} and WebArena~\cite{zhou2023webarena}, success rates degrade sharply as task length grows, with agents forgetting prior observations, repeating actions, or losing track of sub-goals.
The bottleneck has therefore shifted from perception to cross-step state management.
To cope with growing context, prior work restructures history as structured prompts or program variables~\cite{tian2025agentprog,wang2026history}, compresses trajectory tokens~\cite{chen2025less}, or maintains rule-based skill memory for computer control~\cite{tan2024cradle}.
These approaches improve how agents reason over history, but leave open what should be retained, when it should be compressed, and how experience should be reused over time.

\section{Data Curation}
\begin{figure}[htbp]
\vspace{-2mm}
	\centering
	\includegraphics[width=0.92\columnwidth]{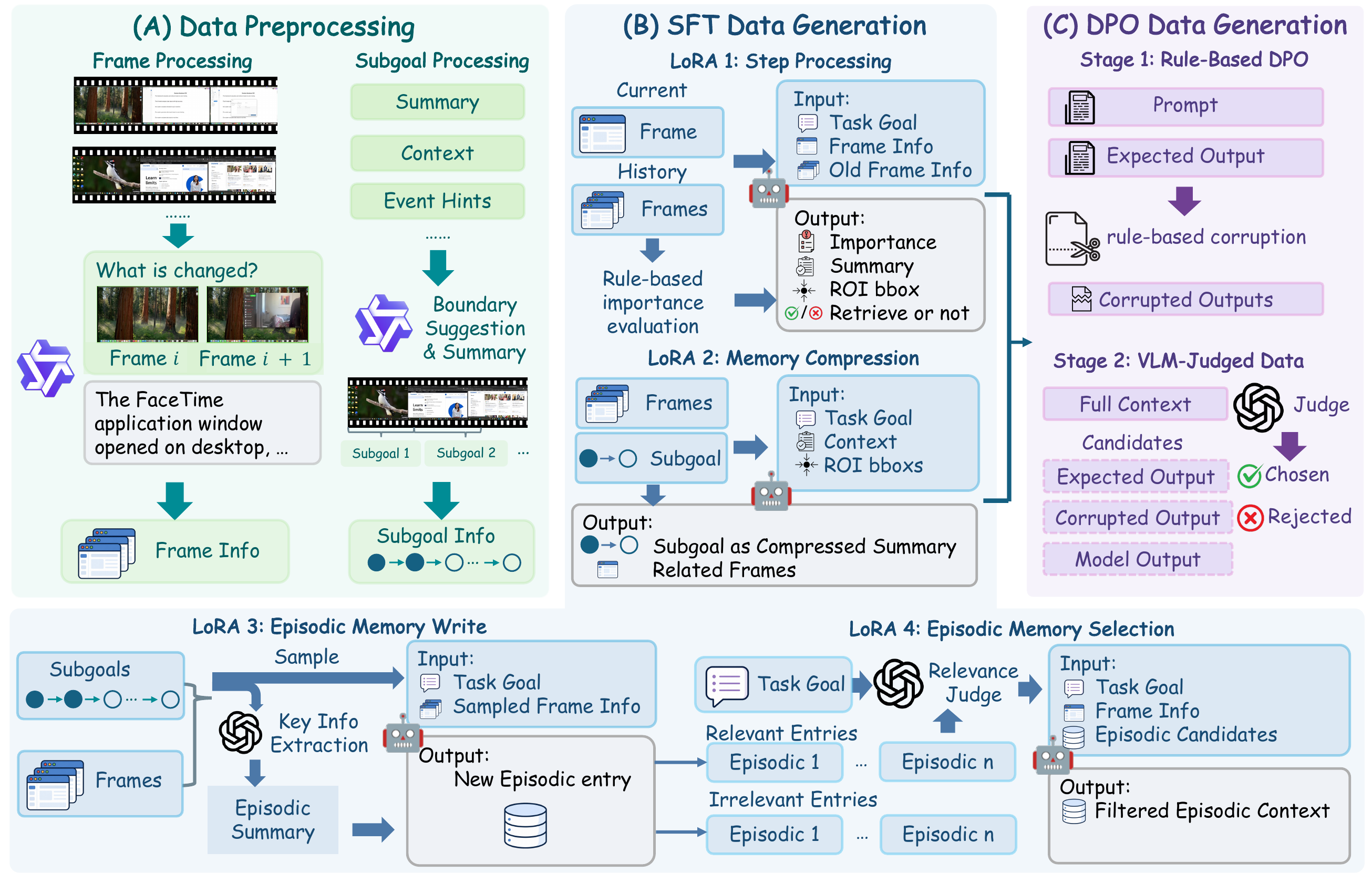}
        \vspace{-2mm}
	\caption{
Overview of the \NAME data curation pipeline.
(A) Raw computer-use videos are parsed into hierarchical frame- and subgoal-level annotations.
(B) These annotations are converted into SFT data for four \SYSTEM operators: step processing, memory compression, episodic memory writing, and episodic memory selection.
(C) Step-processing and memory-compression samples are further corrupted and VLM-filtered to form DPO training pairs.
}
	\label{fig:data_curation}
    \vspace{-4mm}
\end{figure}

To train the memory controller, we curate structured supervision from raw computer-use trajectories in PSAI~\cite{psai_computer_use_2025}. As illustrated in Figure ~\ref{fig:data_curation}, the pipeline first preprocesses the raw video and metadata into frame-level and subgoal-level annotations, then uses the annotations to construct the SFT training data for four memory-control operators. Finally, preference pairs for the online memory operators are constructed through rule-based corruption and VLM-judged filtering. We assess annotation quality through human validation on 200 randomly sampled trajectories, of which 197 are judged fully correct.

\subsection{Data Preprocessing}

Each trajectory in the raw computer-use dataset is converted into two annotation streams. Frame-level annotations capture fine-grained interface transitions by comparing adjacent video frames, including action occurrence, event description, input type, key sequence when applicable, and an ROI box for the changed interface region. Subgoal-level annotations capture coarse task progress by segmenting metadata events and interaction logs into chronological semantic units.

\subsection{Memory Supervision Construction}
\label{sec:memory_supervision_construction}
We convert the preprocessed frame and subgoal annotations into operator-specific supervision for \SYSTEM. Specifically, we construct four supervised datasets, $\mathcal{D}_{\mathrm{step}}$, $\mathcal{D}_{\mathrm{cmp}}$, $\mathcal{D}_{\mathrm{write}}$, and $\mathcal{D}_{\mathrm{sel}}$, corresponding to the Step Processor, WM Compressor, Episodic Writer, and Episodic Selector. Each example pairs the task goal and relevant multimodal context with a structured target following the schema of the corresponding memory operation.

For SFT, step-processing examples are constructed from adjacent-frame annotations and subgoal context, with targets including importance scores, event summaries, ROI bounding boxes, and episodic-retrieval activation tags. Compression examples are built by simulating working-memory buffers and asking the model to summarize older entries while preserving representative visual identifiers. Episodic-writing examples convert completed trajectories into compact reusable memories, and episodic-selection examples train the model to filter retrieved candidates by relevance to the current task state. We further construct DPO preference data for the Step Processor and WM Compressor, the two operators most directly tied to online memory quality. Preference pairs are obtained in two stages: rule-based corruptions create controlled negative outputs, and VLM-judged filtering selects outputs that better preserve task-relevant state, maintain visual grounding, and provide useful downstream context. The resulting preference sets are used for DPO training of the Step Processor and WM Compressor.

\section{Methodology}

\subsection{The \NAME Framework}

Given a task goal $g$ and a long-horizon GUI episode $\mathcal{E}=\{x_t\}_{t=1}^{T}$, where $x_t$ denotes the screenshot at step $t$, the agent predicts actions $\{a_t\}_{t=1}^{T}$ to complete the task. 
We study a plug-in setting where the GUI action model is a frozen backbone $\pi_B$, and \NAME augments it with an external multimodal memory controller, \SYSTEM. 
\SYSTEM implements a deterministic input-construction step and four learned operators: writing salient events into working memory, consolidating older entries, triggering episodic retrieval, and selecting relevant past episodes.

\NAME contains an in-episode working memory $W_t$, a cross-episode episodic memory bank $\mathcal{M}$, and \SYSTEM. 
Working memory tracks transient task state, while episodic memory stores reusable experience from completed episodes. 
\SYSTEM is built by attaching four task-specific LoRA adapters to a shared frozen Qwen3-VL backbone, corresponding to step processing, working-memory compression, episodic writing, and episodic selection. 
Memory exposure is performed by the input constructor, which serializes textual summaries and ROI references into the native multimodal interface of the GUI backbone. 
Thus, \NAME requires no memory-specific tokens, projection layers, architecture changes, or action-backbone finetuning.

\begin{figure}
    \centering
    \includegraphics[width=0.99\columnwidth]{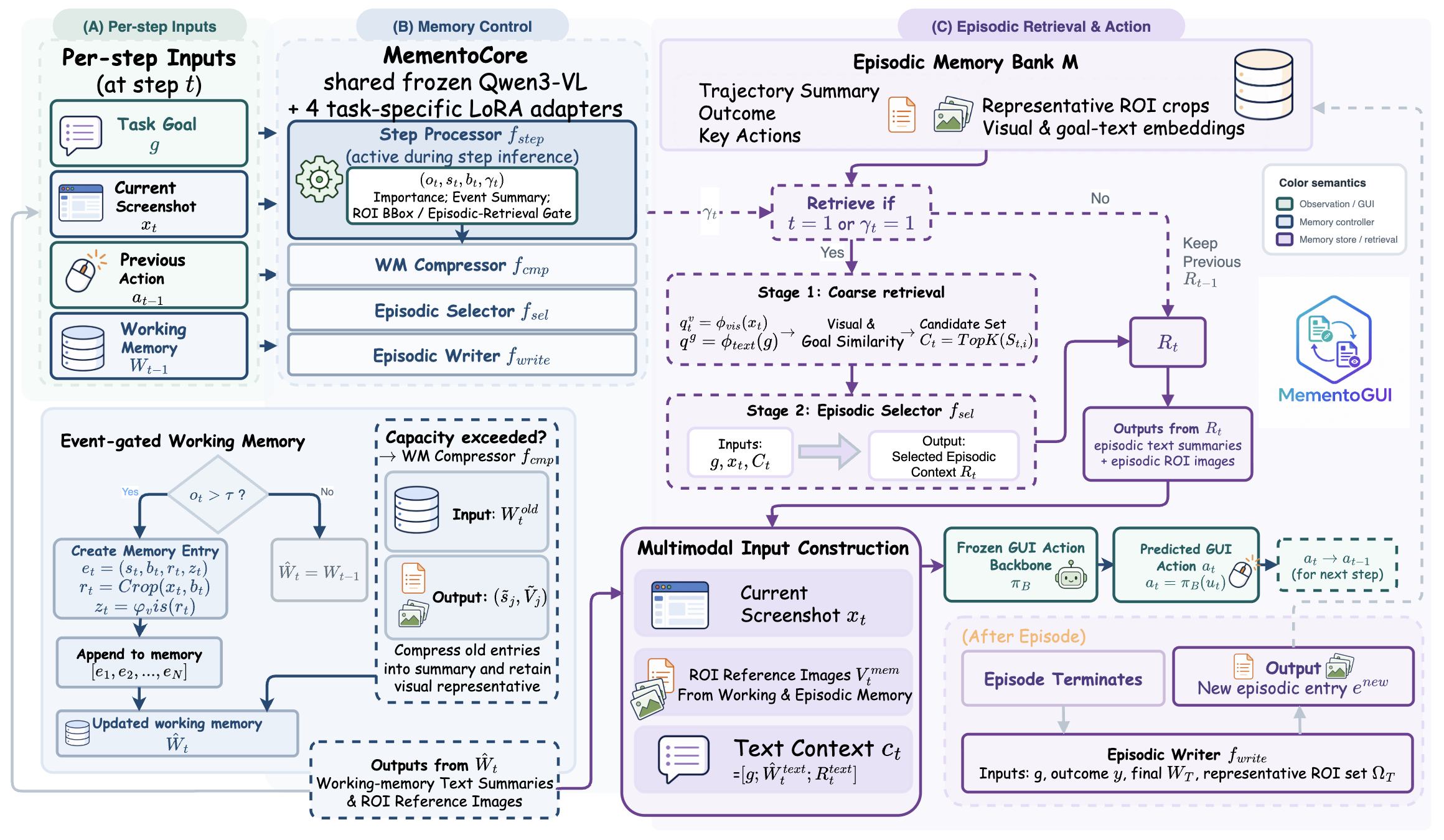}
    \vspace{-2mm}
    \caption{
    \NAME augments a frozen GUI action backbone with multimodal working and episodic memory. 
    It updates, retrieves, and writes memory, then serializes textual summaries and ROI references as multimodal context for GUI action prediction.}
    \label{fig:framework}
    \vspace{-4mm}
\end{figure}

At step $t$, the Step Processor outputs
\begin{equation}
    (o_t, s_t, b_t, \gamma_t)
    =
    f_{\mathrm{step}}(g, x_t, a_{t-1}, W_{t-1}),
    \label{eq:step_processor}
\end{equation}
where $o_t\in[0,1]$ is a write-salience score, $s_t$ is an event summary, $b_t$ is a task-relevant ROI box, and $\gamma_t$ indicates whether episodic retrieval is needed. 
This yields a pre-action working memory $\hat{W}_t$. 
Episodic retrieval is invoked at $t=1$ and, afterward, only when $\gamma_t=1$.

The frozen GUI backbone receives
\begin{equation}
    \mathbf{u}_t = (x_t,\mathcal{V}^{\mathrm{mem}}_t,c_t),
    \qquad
    c_t = [g;\hat{W}^{\mathrm{text}}_t;R_t^{\mathrm{text}}],
    \label{eq:memory_augmented_input}
\end{equation}
where $\mathcal{V}^{\mathrm{mem}}_t$ contains selected ROI images from working and episodic memory, and $c_t$ contains the task goal and textual memory summaries. 
The next action is predicted as
\begin{equation}
    a_t = \pi_B(\mathbf{u}_t),
    \qquad
    W_t \leftarrow \hat{W}_t .
    \label{eq:backbone_action}
\end{equation}
The input $\mathbf{u}_t$ is serialized using the standard multimodal chat template of the backbone, so all memory is consumed as ordinary text and images.

\subsection{Memory System Design}

\subsubsection{Event-Gated Working Memory}
\label{sec:event_gated_wm}

Working memory preserves task-relevant state without replaying the full interaction history. 
Rather than logging every frame, \NAME writes memory only when the current interface may affect future decisions. 
For a retained step, the memory item is
\begin{equation}
    e_t = (s_t, b_t, r_t, z_t), 
    \qquad
    r_t = \mathrm{Crop}(x_t,b_t), 
    \qquad
    z_t = \phi_{\mathrm{vis}}(r_t),
    \label{eq:wm_entry}
\end{equation}
where $r_t$ is the ROI crop and $z_t$ is used only for memory organization. 
The update rule is
\begin{equation}
    \hat{W}_t =
    \begin{cases}
    \mathrm{Append}(W_{t-1}, e_t), & \text{if } o_t > \tau, \\
    W_{t-1}, & \text{otherwise},
    \end{cases}
    \label{eq:wm_update}
\end{equation}
where $\tau$ converts the learned salience score into a deterministic write decision. 
The action backbone never receives $z_t$ as a custom token; selected ROI crops are passed as ordinary images.

To control context growth, older uncompressed entries are consolidated when the recent-memory capacity is exceeded:
\begin{equation}
    (\tilde{s}_j,\tilde{\mathcal{V}}_j)
    =
    f_{\mathrm{cmp}}(g,W_t^{\mathrm{old}}),
    \label{eq:wm_compress}
\end{equation}
where $\tilde{s}_j$ is a compact summary and $\tilde{\mathcal{V}}_j$ contains retained visual identifiers resolved into ROI crops during input construction. 
We pass at most $K_{\mathrm{roi}}$ ROI references to the backbone from compressed blocks and recent entries.

\subsubsection{On-Demand Episodic Memory}
\label{sec:on_demand_em}

Episodic memory stores reusable experience across completed episodes. 
Each entry contains a trajectory summary, metadata such as outcome and key actions, representative ROI crops, and retrieval embeddings. 
Unlike static retrieval, \NAME initializes episodic context at the first step and refreshes it only when $\gamma_t=1$.

When retrieval is invoked, \NAME first performs coarse retrieval using the current screenshot and task goal:
\begin{equation}
    s_{t,i} =
    \lambda_v \cos(q_t^v,m_i^v)
    +
    \lambda_g \cos(q^g,m_i^g),
    \qquad
    \mathcal{C}_t = \mathrm{TopK}_{i}(s_{t,i}),
    \label{eq:em_coarse_retrieval}
\end{equation}
where $q_t^v=\phi_{\mathrm{vis}}(x_t)$, $q^g=\phi_{\mathrm{text}}(g)$, and $(m_i^v,m_i^g)$ are visual and goal-text embeddings of episodic entry $m_i$. 
The Episodic Selector then filters the coarse candidates:
\begin{equation}
    \tilde{R}_t =
    f_{\mathrm{sel}}(g,x_t,\{m_i\}_{i\in\mathcal{C}_t}),
    \label{eq:em_select}
\end{equation}
where each $m_i$ includes its summary, metadata, and ROI crops. 
The episodic context is updated by
\begin{equation}
    R_t =
    \begin{cases}
    \tilde{R}_t, & \text{if } t=1 \text{ or } \gamma_t=1, \\
    R_{t-1}, & \text{otherwise}.
    \end{cases}
    \label{eq:em_update}
\end{equation}
This two-stage design combines efficient vector retrieval with multimodal relevance filtering, while allowing the accumulated working memory to gate when retrieval is invoked.

After an episode ends, the Episodic Writer converts the trajectory into a compact memory:
\begin{equation}
    e^{\mathrm{new}}
    =
    f_{\mathrm{write}}(g,y,W_T,\Omega_T),
    \label{eq:em_write}
\end{equation}
where $y$ is the outcome and $\Omega_T$ is the representative ROI set from the final working memory. 
The new entry is stored in $\mathcal{M}$ with its metadata, embeddings, and ROI crops.

\subsection{Training \SYSTEM}
\label{sec:training_memory_controllers}

We train the four LoRA adapters of \SYSTEM as structured memory-control tasks on top of a shared frozen Qwen3-VL~\cite{bai2025qwen3} backbone. 
The supervised datasets 
$\mathcal{D}_{\mathrm{step}}$, 
$\mathcal{D}_{\mathrm{cmp}}$, 
$\mathcal{D}_{\mathrm{write}}$, and 
$\mathcal{D}_{\mathrm{sel}}$ 
are produced by the data curation pipeline in Section~\ref{sec:memory_supervision_construction}. 
For each operator $k\in\{\mathrm{step},\mathrm{cmp},\mathrm{write},\mathrm{sel}\}$ with LoRA parameters $\alpha_k$ and frozen backbone parameters $\theta$, we minimize
\begin{equation}
    \mathcal{L}_{\mathrm{SFT}}^{(k)}
    =
    -\mathbb{E}_{(x,y)\sim \mathcal{D}_k}
    \log p_{\theta,\alpha_k}(y \mid x).
    \label{eq:sft_memory}
\end{equation}

We further apply DPO to the Step Processor and WM Compressor using preference sets $\mathcal{P}_{\mathrm{step}}$ and $\mathcal{P}_{\mathrm{cmp}}$, since these operators directly trade off informativeness against context budget. 
The Episodic Writer and Selector have direct supervised targets and are trained with SFT only. 
For $k\in\{\mathrm{step},\mathrm{cmp}\}$, DPO is initialized from the SFT adapter $\alpha_k^{\mathrm{SFT}}$, with reference policy $p_{\mathrm{ref},k}=p_{\theta,\alpha_k^{\mathrm{SFT}}}$. Given $(x,y^+,y^-)\sim\mathcal{P}_k$, we optimize
\begin{equation}
\mathcal{L}_{\mathrm{DPO}}^{(k)}
=
-\mathbb{E}
\log \sigma \left[
\beta
\log
\frac{
p_{\theta,\alpha_k}(y^+ \mid x)\,p_{\mathrm{ref},k}(y^- \mid x)
}{
p_{\mathrm{ref},k}(y^+ \mid x)\,p_{\theta,\alpha_k}(y^- \mid x)
}
\right].
\label{eq:dpo_memory}
\end{equation}

\subsection{Benchmarking Long-Horizon GUI Agents}

\paragraph{\DATA.}
We construct \DATA, an offline benchmark derived from PSAI computer-use videos~\cite{psai_computer_use_2025} for memory-dependent GUI decision making. 
It contains 200 trajectories with 6,953 steps, averaging 34.8 steps per trajectory, 80 for testing and 120 for test-time scaling, and focuses on cases where the next action depends on accumulated task state, delayed constraints, completed subgoals, or prior experience. All reported \DATA results are evaluated on the 80 trajectories, and another 120  are used to accumulate episodic memory.

\paragraph{Semantic and Memory-Aware Evaluation Framework.}
Reference-based GUI evaluation is standardized but incomplete for long-horizon tasks, where multiple action paths may be valid and decision quality depends on accumulated state. 
We therefore report VLM-based metrics alongside standard reference-based scores. 
\textbf{VLM-based Action Match (VAM)} measures whether a predicted action is semantically equivalent to the reference action on the current screenshot. 
\textbf{Task Progress Score (TPS)} evaluates whether the predicted sequence moves the task forward without loops, regressions, or stalling. 
\textbf{Memory Consistency Score (MCS)} assesses whether the memory state evolves consistently with task progress, including prior selections, completed subgoals, user constraints, and retrieved episodic experience.
\section{Experiments}
\label{sec:experiment}
\begin{table}[t]
\centering
\caption{
Quantitative results on three GUI benchmarks. 
We evaluate history and memory augmentation strategies on four frozen open-source GUI backbones, with closed-source generalist MLLMs included as direct-prompting baselines. 
Metrics include AMS/Traj. SR for GUI-Odyssey, Step SR for MM-Mind2Web, and VAM/TPS/MCS for MementoGUI-Bench.
}
\label{tab:main_results}
\resizebox{\columnwidth}{!}{%
\begin{tabular}{llcccccc}
\toprule
\multirow{2}{*}{Backbone} 
& \multirow{2}{*}{Method}
& \multicolumn{2}{c}{GUI-Odyssey}
& \multicolumn{1}{c}{MM-Mind2Web}
& \multicolumn{3}{c}{MementoGUI-Bench} \\
\cmidrule(lr){3-4}
\cmidrule(lr){5-5}
\cmidrule(lr){6-8}
& 
& AMS $\uparrow$ & Traj. SR $\uparrow$
& Step SR $\uparrow$
& VAM $\uparrow$ & TPS $\uparrow$ & MCS $\uparrow$ \\
\midrule

\rowcolor[HTML]{e9edf6}
\multicolumn{8}{c}{\textit{Proprietary generalist MLLM agents}} \\
\midrule
GPT-5.5~\cite{singh2025openai}
& Direct Prompting                   
& 54.46 & 2.02
& 18.81
& 1.95 & 2.00 & 2.86 \\
Gemini-3.1-Pro~\cite{googledeepmind2026gemini31pro}
& Direct Prompting                   
& 60.62 & 1.81
& 22.97
& 2.18 & 2.67 & 2.75 \\

\midrule
\rowcolor[HTML]{e9edf6}
\multicolumn{8}{c}{\textit{Open-source frozen GUI backbones}} \\
\midrule

\multirow{5}{*}{UI-Venus-1.5-8B~\cite{gao2026ui}}
& No History                         
& 54.58 & 1.29
& 5.29
& \textbf{1.80} & 2.00 & 0.00 \\
& Pred. Hist. All                    
& 66.31 & 2.33
& 7.57
& 1.58 & 2.29 & 5.36 \\
& Text Summary Memory                
& 62.18 & 2.12
& 11.66
& 1.58 & 3.38 & 3.08 \\
& \memcell{\textbf{Working Memory}}                     
& \memcell{\underline{67.69}} & \memcell{\underline{2.69}}
& \memcell{\underline{11.80}}
& \memcell{1.41} & \memcell{\underline{4.63}} & \memcell{\underline{7.00}} \\
& \memcell{\textbf{Working + Episodic Memory}}          
& \memcell{\textbf{68.32}} & \memcell{\textbf{3.57}}
& \memcell{\textbf{12.60}}
& \memcell{\underline{1.67}} & \memcell{\textbf{5.16}} & \memcell{\textbf{7.14}} \\
\midrule

\multirow{5}{*}{MAI-UI-8B~\cite{zhou2025mai}}
& No History                         
& 35.70 & 0.36
& 12.53
& \underline{1.81} & 2.00 & 0.00 \\
& Pred. Hist. All                    
& 44.79 & 1.35
& 13.44
& 1.54 & 2.11 & 7.55 \\
& Text Summary Memory                
& 38.33 & 0.62
& 13.28
& 1.63 & 3.00 & 3.23 \\
& \memcell{\textbf{Working Memory}}                     
& \memcell{\underline{49.08}} & \memcell{\underline{1.97}}
& \memcell{\underline{14.61}}
& \memcell{1.76} & \memcell{\underline{5.13}} & \memcell{\textbf{8.18}} \\
& \memcell{\textbf{Working + Episodic Memory}}          
& \memcell{\textbf{49.31}} & \memcell{\textbf{2.12}}
& \memcell{\textbf{14.67}}
& \memcell{\textbf{1.88}} & \memcell{\textbf{5.36}} & \memcell{\underline{8.13}} \\
\midrule

\multirow{5}{*}{GUI-Owl-1.5-8B~\cite{xu2026mobile}}
& No History                         
& 40.15 & 0.16
& \textbf{17.78}
& 1.60 & 2.00 & 0.00 \\
& Pred. Hist. All                    
& 38.88 & 0.47
& 13.56
& 1.04 & 2.00 & 5.25 \\
& Text Summary Memory                
& 45.45 & 0.62
& 14.14
& 1.51 & 3.05 & 3.10 \\
& \memcell{\textbf{Working Memory}}                     
& \memcell{\underline{48.25}} & \memcell{\underline{1.40}}
& \memcell{15.53}
& \memcell{\underline{1.79}} & \memcell{\underline{3.38}} & \memcell{\underline{7.39}} \\
& \memcell{\textbf{Working + Episodic Memory}}          
& \memcell{\textbf{49.45}} & \memcell{\textbf{1.71}}
& \memcell{\underline{16.42}}
& \memcell{\textbf{1.82}} & \memcell{\textbf{4.18}} & \memcell{\textbf{7.73}} \\
\midrule

\multirow{5}{*}{GUI-Owl-1.5-32B~\cite{xu2026mobile}}
& No History                         
& 45.73 & 0.57
& \underline{18.98}
& 2.10 & 2.00 & 0.00 \\
& Pred. Hist. All                    
& 49.02 & 1.55
& 16.48
& 1.83 & 2.11 & 4.73 \\
& Text Summary Memory                
& 47.21 & 0.72
& 17.00
& 2.37 & 3.40 & 3.40 \\
& \memcell{\textbf{Working Memory}}            
& \memcell{\underline{51.36}} & \memcell{\underline{2.33}}
& \memcell{17.71}
& \memcell{\underline{2.50}} & \memcell{\underline{4.24}} & \memcell{\underline{7.85}} \\
& \memcell{\textbf{Working + Episodic Memory}} 
& \memcell{\textbf{55.17}} & \memcell{\textbf{2.59}}
& \memcell{\textbf{19.12}}
& \memcell{\textbf{2.89}} & \memcell{\textbf{4.31}} & \memcell{\textbf{8.30}} \\

\bottomrule
\end{tabular}%
}
\vspace{-5mm}
\end{table}
\subsection{Experimental Setup}
\paragraph{Implementation Details.}
All GUI backbones are frozen during evaluation. 
We evaluate four open-source GUI models: UI-Venus-1.5-8B~\cite{gao2026ui}, MAI-UI-8B~\cite{zhou2025mai}, GUI-Owl-1.5-8B, and GUI-Owl-1.5-32B~\cite{xu2026mobile}. 
\NAME is used as a plug-in memory controller that injects working- and episodic-memory context into the backbone prompt without backbone finetuning, using controllers trained as described in Sec.~\ref{sec:training_memory_controllers}. 
We also evaluate GPT-5.5~\cite{singh2025openai} and Gemini-3.1-Pro~\cite{googledeepmind2026gemini31pro} as API-based MLLM agents under the same observation, instruction, and action format. Latency is measured on NVIDIA H100 GPUs under the same inference setup and includes memory-controller inference, retrieval, prompt construction, and GUI-backbone inference.

\paragraph{Evaluation Metrics.}
We report standard offline GUI metrics following each benchmark protocol, including action-matching score (AMS), step success rate, and trajectory-level success rate. 
For finer-grained analysis, we additionally use VAM, TPS, and MCS to measure semantic action matching, task-progress plausibility, and consistency with accumulated task state/memory context, respectively, with Gemini-3.1-Pro as the VLM judge. 
For API-based and scaling experiments, we further report trajectory-level inference time and token usage to quantify compute and context cost.

\begin{figure}[htbp]
    \centering
    \includegraphics[width=\columnwidth]{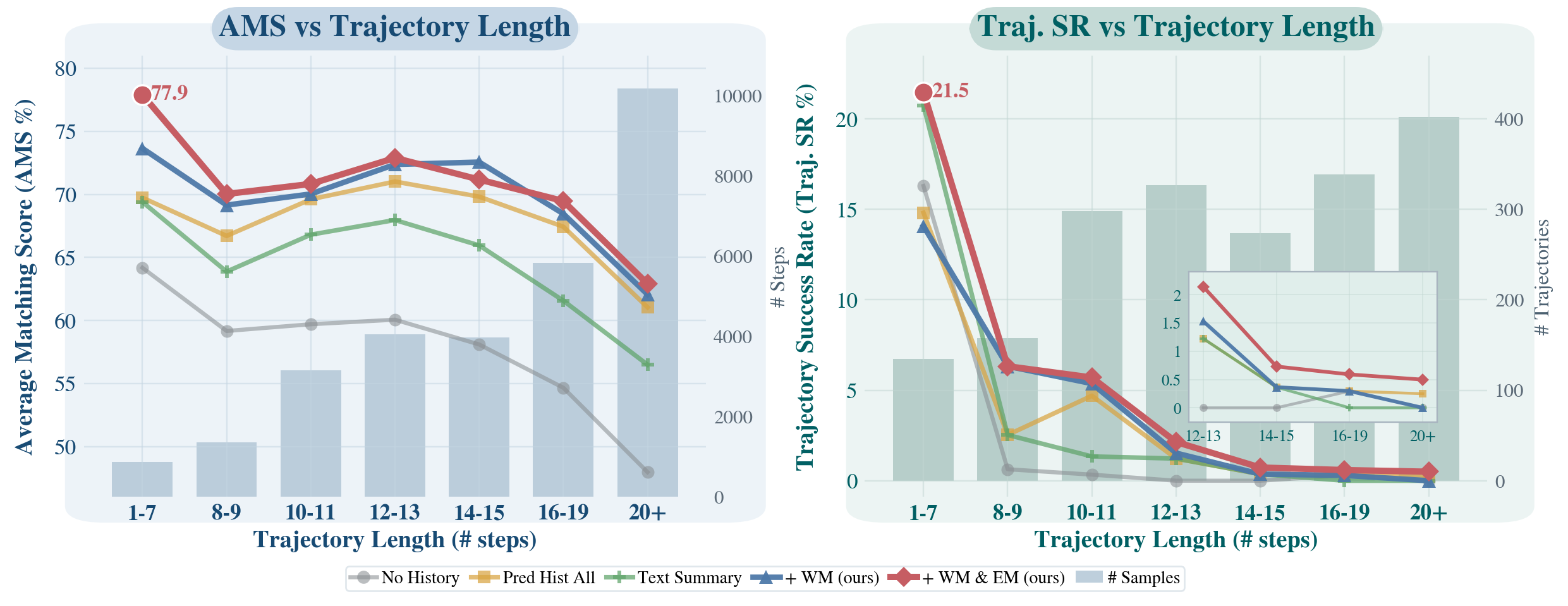}
    \caption{GUI-Odyssey performance by trajectory length on UI-Venus-1.5-8B.}
    \label{fig:method_compare}
\vspace{-4mm}
\end{figure}

\subsection{Quantitative Results}
\begin{figure}[htbp]
    \centering
    \includegraphics[width=\columnwidth]{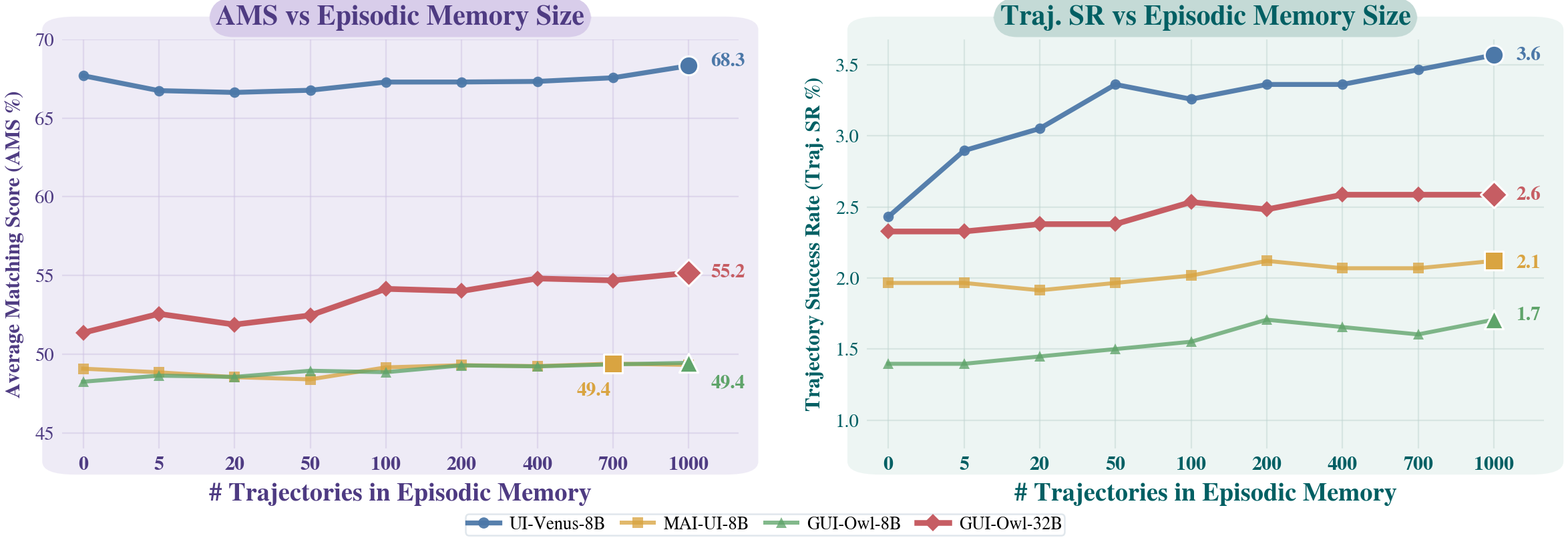}
    \caption{Effect of episodic memory bank size on GUI-Odyssey across frozen GUI backbones.}
    \label{fig:em_size}
\vspace{-4mm}
\end{figure}

Table~\ref{tab:main_results} reports results on GUI-Odyssey~\cite{lu2025guiodyssey}, MM-Mind2Web~\cite{deng2023mind2web}, and \DATA. 
\NAME consistently improves frozen open-source GUI backbones over no-history, predicted-history, and text-summary baselines. 
For example, on GUI-Odyssey with UI-Venus-1.5-8B, working memory raises AMS from 54.58 to 67.69 and trajectory success from 1.29 to 2.69; adding episodic memory further improves them to 68.32 and 3.57. 
Similar gains across MAI-UI-8B and GUI-Owl variants confirm its effectiveness as plug-in memory augmentation. 
Figure~\ref{fig:method_compare} shows stronger AMS across trajectory-length bins and higher trajectory success than history-based and text-only memory baselines, especially when working and episodic memory are combined. 
Figure~\ref{fig:em_size} shows that larger episodic memory banks generally improve trajectory success, suggesting that reusable experience mainly benefits long-horizon completion. 
Table~\ref{tab:api_wm} further shows that the same working-memory context can augment proprietary MLLM agents in a stateless single-step setting.

\begin{table}[htbp]
\centering
\caption{
Plug-in working-memory augmentation for API-based MLLM agents. 
Working Memory replaces native conversation history with stateless single-step inference conditioned on the current screenshot, task instruction, and MementoGUI memory context.
}
\vspace{-2mm}
\label{tab:api_wm}
\resizebox{\columnwidth}{!}{%
\begin{tabular}{llllcclllcc}

\toprule
\multirow{2}{*}{Backbone}
& \multirow{2}{*}{Method}
& \multicolumn{4}{c}{GUI-Odyssey}
& \multicolumn{5}{c}{MementoGUI-Bench} \\
\cmidrule(lr){3-6}
\cmidrule(lr){7-11}
&
& AMS $\uparrow$ & Traj. SR $\uparrow$
& Token & Time
& VAM $\uparrow$ & TPS $\uparrow$ & MCS $\uparrow$
& Token & Time \\
\midrule

\multirow{2}{*}{GPT-5.5}
& Direct Prompting
& 54.46 & 2.02
& 29,974 & 66.28 s
& 1.95 & 2.00 & 2.86
& 80,066 & 133.15 s \\
& \memcell{\textbf{Working Memory}}
& \memcell{55.73\textcolor{my_green}{$_{+2.33\%}$}} 
& \memcell{2.07\textcolor{my_green}{$_{+2.48\%}$}}
& \memcell{37,798} & \memcell{80.73 s}
& \memcell{1.63\textcolor{my_red}{$_{-16.41\%}$}} 
& \memcell{2.57\textcolor{my_green}{$_{+28.50\%}$}} 
& \memcell{6.57\textcolor{my_green}{$_{+129.72\%}$}}
& \memcell{66,718} & \memcell{174.69 s} \\
\midrule

\multirow{2}{*}{Gemini-3.1-Pro}
& Direct Prompting
& 60.62 & 1.81
& 26,100 & 58.44 s
& 2.18  & 2.67 & 2.75
& 75,769 & 125.37 s \\
& \memcell{\textbf{Working Memory}}
& \memcell{61.94\textcolor{my_green}{$_{+2.18\%}$}} 
& \memcell{1.97\textcolor{my_green}{$_{+8.84\%}$}}
& \memcell{43,918} & \memcell{78.91 s}
& \memcell{2.59\textcolor{my_green}{$_{+18.81\%}$}} 
& \memcell{3.71\textcolor{my_green}{$_{+38.95\%}$}} 
& \memcell{7.22\textcolor{my_green}{$_{+162.55\%}$}}
& \memcell{106,237} & \memcell{160.82 s} \\

\bottomrule
\end{tabular}
}
\vspace{-4mm}
\end{table}

\subsection{Ablation Study}
Table~\ref{tab:wm_visual_ablation} studies whether working-memory gains come only from learned text summarization or also require visual grounding. \textit{WM w/o Visual Memory} uses the same learned memory-writing and compression controller as full working memory but removes ROI reference images from the memory context, isolating the contribution of localized GUI visual evidence. Table~\ref{tab:em_retrieval_ablation} studies how episodic memories should be selected before injection. \textit{Random Episodic Context} controls for adding unrelated past experience, \textit{Single-stage Retrieval} directly uses the top embedding-retrieved episode, and \textit{Two-stage Retrieval} corresponds to our full WM+EM setting with learned relevance selection. Across both backbones, removing visual memory or learned episodic selection consistently degrades performance, confirming that \NAME benefits from both ROI-level grounding and filtered episodic experience rather than merely adding more context.

\begin{table}[htbp]
\centering
\scriptsize
\vspace{-2mm}
\begin{minipage}[htbp]{0.49\columnwidth}
\centering
\caption{
Ablation of visual grounding in working memory. 
\textit{WM w/o Visual Memory} refers to using the learned memory control but removing ROI reference images from the memory context, while full \textit{Working Memory} uses both textual summaries and selected ROI images.
}
\label{tab:wm_visual_ablation}

\resizebox{\linewidth}{!}{%
\begin{tabular}{lccccc}
\toprule
\multirow{2}{*}{Method}
& \multicolumn{2}{c}{GUI-Odyssey}
& \multicolumn{3}{c}{MementoGUI-Bench} \\
\cmidrule(lr){2-3}
\cmidrule(lr){4-6}
& AMS $\uparrow$ & Traj. SR $\uparrow$
& VAM $\uparrow$ & TPS $\uparrow$ & MCS $\uparrow$ \\
\midrule

\rowcolor[HTML]{e9edf6}
\multicolumn{6}{c}{Backbone: UI-Venus-1.5-8B~\cite{gao2026ui}} \\
\midrule
Text Summary Memory
& 62.18 & 2.12
& \textbf{1.58} & 3.38 & 3.08 \\
WM w/o Visual Memory
& 64.22 & 2.48
& 1.53 & 4.50 & 6.68\\
\memcell{\textbf{Working Memory}}
& \memcell{\textbf{67.69}} & \memcell{\textbf{2.69}}
& \memcell{1.41} & \memcell{\textbf{4.63}} & \memcell{\textbf{7.00}} \\

\midrule
\rowcolor[HTML]{e9edf6}
\multicolumn{6}{c}{Backbone: GUI-Owl-1.5-8B~\cite{xu2026mobile}} \\
\midrule
Text Summary Memory
& 45.45 & 0.62
& 1.51 & 3.05 & 3.10 \\
WM w/o Visual Memory
& 47.58 & 1.19
& 1.76 & 3.33 & 5.78  \\
\memcell{\textbf{Working Memory}}
& \memcell{\textbf{48.25}} & \memcell{\textbf{1.40}}
& \memcell{\textbf{1.79}} & \memcell{\textbf{3.38}} & \memcell{\textbf{7.39}} \\

\bottomrule
\end{tabular}%
}
\end{minipage}
\vspace{-2mm}
\hfill
\begin{minipage}[htbp]{0.49\columnwidth}
\centering
\caption{
Ablation of episodic retrieval. 
\textit{Single-stage Retrieval} injects the top embedding-retrieved episode, while \textit{Two-stage Retrieval} applies learned relevance selection before memory injection.
}
\label{tab:em_retrieval_ablation}

\resizebox{\linewidth}{!}{%
\begin{tabular}{lccccc}
\toprule
\multirow{2}{*}{Method}
& \multicolumn{2}{c}{GUI-Odyssey}
& \multicolumn{3}{c}{MementoGUI-Bench} \\
\cmidrule(lr){2-3}
\cmidrule(lr){4-6}
& AMS $\uparrow$ & Traj. SR $\uparrow$
& VAM $\uparrow$ & TPS $\uparrow$ & MCS $\uparrow$ \\
\midrule

\rowcolor[HTML]{e9edf6}
\multicolumn{6}{c}{Backbone: UI-Venus-1.5-8B~\cite{gao2026ui}} \\
\midrule
WM Only
& 67.69 & 2.69
& 1.41 & 4.63 & 7.00 \\
Random Episodic Context
& 64.46 & 2.22
& 1.31 & 4.48 & 6.35 \\
Single-stage Retrieval
& 64.40 & 2.33
& 1.62 & 4.79 & 6.65 \\
\memcell{\textbf{Two-stage Retrieval}}
& \memcell{\textbf{68.32}} & \memcell{\textbf{3.57}}
& \memcell{\textbf{1.67}} & \memcell{\textbf{5.16}} & \memcell{\textbf{7.14}} \\

\midrule
\rowcolor[HTML]{e9edf6}
\multicolumn{6}{c}{Backbone: GUI-Owl-1.5-8B~\cite{xu2026mobile}} \\
\midrule
WM Only
& 48.25 & 1.40
& 1.79 & 3.38 & 7.39 \\
Random Episodic Context
& 45.66 & 1.29
& 1.72 & 3.48 & 6.83 \\
Single-stage Retrieval
& 48.08 & 1.19
& 1.69 & 3.89 & 7.51 \\
\memcell{\textbf{Two-stage Retrieval}}
& \memcell{\textbf{49.45}} & \memcell{\textbf{1.71}}
& \memcell{\textbf{1.82}} & \memcell{\textbf{4.18}} & \memcell{\textbf{7.73}} \\

\bottomrule
\end{tabular}%
}
\end{minipage}
\vspace{-2mm}
\end{table} 

\subsection{Scaling \SYSTEM}

Table~\ref{tab:mementogui_scaling} studies the effect of memory-controller scale while keeping the memory architecture and frozen GUI action backbone fixed. 
We compare 2B, 4B, and 8B \SYSTEM variants under working-memory-only and working-plus-episodic-memory settings and report both GUI performance and end-to-end trajectory latency. Increasing controller capacity generally improves long-horizon decision support, especially in the working-plus-episodic-memory setting. 
The 8B controller achieves the strongest results on several key metrics, including action matching for both UI-Venus-1.5-8B and GUI-Owl-1.5-8B in GUI-Odyssey, as well as VAM for GUI-Owl-1.5-8B in \DATA. 
Episodic memory further provides complementary gains over working memory alone in most settings, suggesting that retrieved past evidence improves decisions beyond compressed in-task state. 
These improvements come with additional latency in some cases but require no finetuning of the underlying action model. 
Overall, the results indicate that \NAME can scale as a plug-in memory layer by replacing the controller with stronger variants.

\begin{table}[htbp]
\vspace{-2mm}
\centering
\caption{
We vary the memory-controller size with fixed memory architecture and frozen GUI backbones, comparing working memory with working-plus-episodic memory.
Results report GUI performance and end-to-end trajectory latency on GUI-Odyssey and \DATA.
}
\vspace{-2mm}
\label{tab:mementogui_scaling}
\resizebox{\columnwidth}{!}{%
\begin{tabular}{lllllclllc}
\toprule
\multirow{2}{*}{Backbone}
& \multirow{2}{*}{\#Params}
& \multirow{2}{*}{Method}
& \multicolumn{3}{c}{GUI-Odyssey}
& \multicolumn{4}{c}{MementoGUI-Bench} \\
\cmidrule(lr){4-6}
\cmidrule(lr){7-10}
& & 
& AMS $\uparrow$ & Traj. SR $\uparrow$ & Time
& VAM $\uparrow$ & TPS $\uparrow$ & MCS $\uparrow$ & Time  \\
\midrule

\multirow{6}{*}{UI-Venus-1.5-8B~\cite{gao2026ui}}
& \multirow{2}{*}{2B}
& Working Memory
& 67.69 & 2.69 & 41.98 s
& 1.41 & 4.63 & 7.00 & 86.91 s \\
&
& \memcell{\textbf{Working + Episodic Memory}}
& \memcell{68.32\textcolor{my_green}{$_{+0.93\%}$}} 
& \memcell{3.57\textcolor{my_green}{$_{+32.71\%}$}} 
& \memcell{51.68 s}
& \memcell{1.67\textcolor{my_green}{$_{+18.44\%}$}} 
& \memcell{5.16\textcolor{my_green}{$_{+11.45\%}$}} 
& \memcell{7.14\textcolor{my_green}{$_{+2.00\%}$}} 
& \memcell{100.84 s} \\
\cmidrule(lr){2-10}

& \multirow{2}{*}{4B}
& Working Memory
& 67.82 & 2.74 & 61.95 s
& 1.53 & 4.84 & 7.36 & 124.60 s \\
&
& \memcell{\textbf{Working + Episodic Memory}}
& \memcell{68.84\textcolor{my_green}{$_{+1.50\%}$}} 
& \memcell{3.20\textcolor{my_green}{$_{+16.79\%}$}} 
& \memcell{60.76 s}
& \memcell{1.44\textcolor{my_red}{$_{-5.88\%}$}} 
& \memcell{4.90\textcolor{my_green}{$_{+1.24\%}$}} 
& \memcell{7.71\textcolor{my_green}{$_{+4.76\%}$}} 
& \memcell{140.63 s} \\
\cmidrule(lr){2-10}

& \multirow{2}{*}{8B}
& Working Memory
& 67.38 & 2.85 & 72.47 s
& 1.60 & 4.83 & 7.39 & 105.52 s \\
&
& \memcell{\textbf{Working + Episodic Memory}}
& \memcell{69.83\textcolor{my_green}{$_{+3.64\%}$}} 
& \memcell{3.36\textcolor{my_green}{$_{+17.89\%}$}} 
& \memcell{56.20 s}
& \memcell{1.63\textcolor{my_green}{$_{+1.87\%}$}} 
& \memcell{4.96\textcolor{my_green}{$_{+2.69\%}$}} 
& \memcell{7.96\textcolor{my_green}{$_{+7.71\%}$}} 
& \memcell{124.78 s} \\
\midrule

\multirow{6}{*}{GUI-Owl-1.5-8B~\cite{xu2026mobile}}
& \multirow{2}{*}{2B}
& Working Memory
& 48.25 & 1.40 & 49.02 s
& 1.79 & 3.38 & 7.39 & 83.28 s \\
&
& \memcell{\textbf{Working + Episodic Memory}}
& \memcell{49.45\textcolor{my_green}{$_{+2.49\%}$}} 
& \memcell{1.71\textcolor{my_green}{$_{+22.14\%}$}} 
& \memcell{54.57 s}
& \memcell{1.82\textcolor{my_green}{$_{+1.68\%}$}} 
& \memcell{4.18\textcolor{my_green}{$_{+23.67\%}$}} 
& \memcell{7.73\textcolor{my_green}{$_{+4.60\%}$}} 
& \memcell{94.89 s} \\
\cmidrule(lr){2-10}

& \multirow{2}{*}{4B}
& Working Memory
& 48.49 & 1.60 & 60.81 s
& 1.64 & 3.90 & 7.58 & 114.37 s \\
&
& \memcell{\textbf{Working + Episodic Memory}}
& \memcell{49.05\textcolor{my_green}{$_{+1.15\%}$}} 
& \memcell{1.71\textcolor{my_green}{$_{+6.88\%}$}} 
& \memcell{61.21 s}
& \memcell{1.87\textcolor{my_green}{$_{+14.02\%}$}} 
& \memcell{3.68\textcolor{my_red}{$_{-5.64\%}$}} 
& \memcell{7.53\textcolor{my_red}{$_{-0.66\%}$}} 
& \memcell{114.96 s} \\
\cmidrule(lr){2-10}

& \multirow{2}{*}{8B}
& Working Memory
& 49.47 & 1.81 & 64.45 s
& 1.65 & 4.14 & 7.50 & 94.49 s \\
&
& \memcell{\textbf{Working + Episodic Memory}}
& \memcell{50.70\textcolor{my_green}{$_{+2.49\%}$}} 
& \memcell{1.81\textcolor{my_gray}{$_{+0.00\%}$}} 
& \memcell{55.24 s}
& \memcell{2.18\textcolor{my_green}{$_{+32.12\%}$}} 
& \memcell{4.58\textcolor{my_green}{$_{+10.63\%}$}} 
& \memcell{7.75\textcolor{my_green}{$_{+3.33\%}$}} 
& \memcell{107.15 s} \\

\bottomrule
\end{tabular}%
}
\vspace{-6mm}
\end{table}

\section{Conclusion}
\vspace{-2mm}
We introduced \NAME, a plug-in online multimodal memory-control framework for long-horizon GUI agents. Instead of relying on raw history replay or longer context windows, \NAME reframes long-horizon GUI control as active memory control, enabling frozen GUI backbones to selectively update, preserve, compress, and retrieve decision-relevant multimodal state across interface transitions. We further developed an automatic data curation pipeline from PSAI computer-use trajectories and introduced \DATA, a benchmark for memory-dependent long-horizon GUI decision making. Across mobile and web environments, \NAME consistently improves frozen GUI backbones over strong history- and memory-based baselines. Ablations show that localized visual evidence and learned episodic selection provide complementary gains, and scaling experiments suggest that stronger memory controllers can further improve long-horizon decision support. These results suggest that agentic multimodal memory control offers a scalable path toward GUI agents that remain coherent, efficient, and reliable over extended interaction trajectories.

\bibliographystyle{plain}
\bibliography{main}

\end{document}